\documentclass[sigconf,balance=false, table]{acmart}
\usepackage{popets}
\usepackage{graphicx}
\usepackage{textcomp}
\usepackage{booktabs}
\usepackage{setspace}
\usepackage{blindtext}
\usepackage{titlesec}
\usepackage{algorithm}
\usepackage{algcompatible}
\usepackage{mathtools}
\usepackage{bm}
\usepackage{mwe}
\usepackage{amsmath,amsfonts, mathrsfs}
\usepackage{bbm, dsfont}
\usepackage{algpseudocode}
\usepackage{multicol}
\usepackage{multirow}
\usepackage{subcaption}
\usepackage{esvect}

\usepackage{xcolor}
\setcopyright{popets}
\copyrightyear{YYYY}

\acmYear{YYYY}
\acmVolume{YYYY}
\acmNumber{X}
\acmDOI{XXXXXXX.XXXXXXX}
\acmISBN{}
\acmConference{Proceedings on Privacy Enhancing Technologies}
\begin{document}

\title{Second-Order Information Matters: Revisiting Machine Unlearning for Large Language Models}


\author{Kang Gu}
\affiliation{
\institution{Dartmouth College}
\country{USA}
 } 

\author{Md Rafi Ur Rashid}
\affiliation{
\institution{Penn State}
\country{USA}
}

\author{Najrin Sultana}
\affiliation{
\institution{Penn State}
\country{USA}
}
\author{Shagufta Mehnaz}
\affiliation{
\institution{Penn State}
\country{USA}
}


\renewcommand{\shortauthors}{Kang et al.}

\begin{abstract}
With the rapid development of Large Language Models (LLMs), we have witnessed intense competition among the major LLM products like ChatGPT, LLaMa, and Gemini. However, various issues (e.g. privacy leakage and copyright violation) of the training corpus still remain underexplored. For example, the Times sued OpenAI and Microsoft for infringing on its copyrights by using millions of its articles for training. 
From the perspective of LLM practitioners, handling such unintended privacy violations can be challenging. Previous work addressed the ``unlearning" problem of LLMs using gradient information, while they mostly introduced significant overheads like data preprocessing or lacked robustness. In this paper, contrasting with the methods based on first-order information, we revisit the unlearning problem via the perspective of second-order information (Hessian). 
Our unlearning algorithms, which are inspired by classic Newton update, are not only data-agnostic/model-agnostic but also proven to be robust in terms of utility preservation or privacy guarantee. Through a comprehensive evaluation with four NLP datasets as well as a case study on real-world datasets, our methods consistently show superiority over the first-order methods.

\end{abstract}


\maketitle

\section{Introduction}
Recent years have witnessed the prosperity of commercial products powered by large language models (LLMs).  
From the breakthrough of ChatGPT \cite{liu2023summary} to the rise of Midjourney \cite{borji2022generated}, the capacity of LLMs has gone beyond typical text-related tasks such as question answering and code generation. However, ensuring responsible usage of LLM services is of utmost importance, given the lessons of previous privacy violations \cite{Iruda, GPT-3}. In 2021, the first
legal case concerning
an AI chatbot happened in South Korea, where the chatbot \textit{Iruda} violated the Personal Information Protection Act after generating the exact home addresses and bank account numbers of actual individuals unintentionally. Very recently, the Times sued OpenAI and Microsoft, accusing the companies of infringing on its copyrights by using millions of its articles to train ChatGPT \cite{Times-sue-OpenAI}.

To ensure data safety,
data privacy regulations such as GDPR \cite{DGPR} have granted users the right to revoke the use of their data by commercial services. 
However, from the perspective of service providers, making a trained model forget about the knowledge of specific training samples can be much more challenging than just deleting the user data from the database. 
In this paper, we study the problem of how a service provider handles incoming user requests regarding the removal of their data during the life cycle of the LLM service. Retraining the model from scratch ensures the erasure of target samples, while it is extremely expensive to practice for LLMs. It raises the question of how LLM practitioners can unlearn the models with much less time and computational resources. 

Recently, efficient unlearning approaches using gradient information have been explored \cite{jang2022knowledge, eldan2023s, chen-yang-2023-unlearn}. 
Eldan \textit{et al.} studied a novel approach to replace sensitive tokens within target samples with generic counterparts and finetune the LLMs on the modified samples \cite{eldan2023s}. It was shown that they effectively weakened the ability of Llama2 7B model to generate or recall Harry Potter-related content.
Nevertheless, this approach does not generalize easily to non-fiction data since identifying private information is non-trivial and the standard of privacy differs by each individual  \cite{brown2022does}.
Jang \textit{et al.} showed that updating the model parameters by inverting the direction of gradients (gradient ascent) can erase the knowledge of target samples \cite{jang2022knowledge}.
However, since the gradient ascent neglects the rest of the training data, it is uncontrollable how well the knowledge of the remaining data is maintained. The performance of the final model can vary drastically with different random seeds, which leaves the utility questionable. Chen and Yang \cite{chen-yang-2023-unlearn} proposed an efficient unlearning approach by plugging lightweight unlearning layers into transformers. During training, the unlearning layers are learned to forget the requested data while the original LLM layers remained unchanged. Despite its effectiveness, this method modifies the architecture of the original models and has only been evaluated on T5 models. \\

\noindent \textbf{Challenges.}
The tremendous size of parameters as well as the high non-convexity make it challenging to design unlearning algorithms for LLMs. 
 So far, existing LLM unlearning methods either: 1) compromise the generalizability in exchange for better performance on specific data/models \cite{eldan2023s, chen-yang-2023-unlearn}, or 2) run efficiently while lacking robustness \cite{jang2022knowledge}. However, an ideal unlearning method should not only avoid introducing overhead (e.g. data engineering and architecture engineering) but also demonstrate robustness with respect to the effectiveness of erasure or the preservation of model utility. 
 \\

\noindent \textbf{Motivation.}
To this end, we explore novel unlearning strategies for LLMs that
can satisfy the aforementioned properties.
For shallow (convex) machine learning models, certified removal \cite{guo2019certified} can largely remove the influence of deleted data points and derive an upper bound of residual information.
Although certified removal cannot be applied directly to DNNs due to their non-convexity, it still implies the crucial role of second-order information in retaining the knowledge of remaining data and stabilizing the unlearning process.
Intuitively, the model produced by an optimal unlearning algorithm should be indistinguishable from another model that has not seen the unlearning subset, which can be measured by Shannon Information, i.e. Kullback-Leibler (KL) divergence. Golatkar \textit{et al.}
\cite{golatkar2020eternal}
showed that employing Newton update as the unlearning algorithm is sufficient to make the aforementioned KL divergence converge to zero.\\

\noindent \textbf{Our Work.} 
We first show that second-order information (Hessian) of LLMs can be approximated efficiently.
Since the complexity of analytically calculating Hessian is quadratic with respect to the parameter size.
Although it is feasible to do so for shallow models (e.g. logistic regression) \cite{guo2019certified}, it quickly scales up to be prohibitively expensive when it comes to LLMs. Various approximation methods have been studied for accelerating the computation of the Hessian \cite{warnecke2021machine, kurtic2022optimal, golatkar2020eternal}.
Kurtic \textit{et al.} \cite{kurtic2022optimal} proved that inverse empirical Fisher estimation can be accurate and scalable to the dimensionality of BERT models for pruning tasks. 
We adapt the inverse empirical Fisher estimation to multi-gpu setting such that it supports the scale of OPT 2.7B \cite{zhang2022opt} as well as even larger models. Our empirical study shows that second-order information can be accurately approximated for unlearning.

 Thus, we propose two unlearning algorithms for LLMs, namely \textit{Fisher Removal} and \textit{Fisher Forgetting}, which are both derived from Newton update. Compared with gradient ascent, Fisher Removal provides a stronger guarantee for the erasure of the unlearning subset while maintaining the LLM utility at a higher level. However, they both update the model parameters in an aggressive way such that the utility might degenerate noticeably after a few iterations of unlearning. To provide the utility guarantee, we further introduce Fisher Forgetting, a variant of Fisher Removal, which maintains the accuracy of LLMs even after going through multiple unlearning cycles.\\ 

\noindent \textbf{Contributions}
We present two versions of unlearning algorithms for LLMs, which are derived from Newton update, namely \textit{Fisher Removal} and \textit{Fisher Forgetting}. 
An empirical study on four widely used NLP datasets is conducted to compare our methods with other baselines including finetuning, retraining, and gradient ascent.
We further explore their mitigation effects for unintended memorization on two real-world datasets. Moreover, we uncover the relationship between unlearning and DP-SGD \cite{abadi2016deep} through the study of the privacy-utility trade-off.
The contributions of our work are summarized as follows:
\begin{enumerate}
    \item We introduce two novel unlearning strategies for LLMs and show that second-order information plays an important role in achieving robust unlearning outcomes.

    \item We extensively evaluate the capacity of each unlearning approach on four widely used NLP datasets as well as two real-world datasets.
    The codebase will be released to enable reproducibility.

    \item We discover that DP-SGD does not guarantee an equally optimal/suboptimal
trade-off across different datasets, which implies it cannot replace the role of unlearning.
  \item  We discuss the limitations such as the cost of approximating Hessian, and also point out directions for future work.
    
\end{enumerate}

\section{Preliminary}
We first revisit the notion of unlearning from the perspective of KL divergence. Then we further discuss unlearning strategies for convex and non-convex models respectively.
The taxonomy is defined in Table \ref{tab:para_def}.

\begin{table}[h]
    \centering
    \caption{Parameter Definition}
    \begin{tabular}{|c|c|}
    \hline
       Name  & Definition  \\
       \hline
       
       $D^-$  & data samples to be removed \\
       \hline
       $D^+$  & data samples left\\
       \hline
       $D$ & full training data $D = D^+ \bigcup D^-$\\
       \hline
       $\theta$ & model parameters \\
       \hline
       $l$ & coefficient for gradient ascent  \\
       \hline
       $\lambda$  & parameter for initializing Empirical Fisher\\ 
       \hline
       $m$ & the number of recursions for Empirical Fisher \\ \hline
       $\gamma$ & coefficient for Fisher Removal \\
       \hline
       $\mu$ & parameter for Fisher Forgetting \\
       \hline
       $\sigma$ & parameter for Fisher Forgetting\\
       \hline
    \end{tabular}
    
    \label{tab:para_def}
\end{table}

\subsection{Unlearning for General Models}

Let $\theta$ denote the parameters of a model trained on $D = D^- \cup D^+$ and $S(\theta, D)$ be an unlearning function that modifies the parameters such that 
$\theta'= S(\theta, D)$. The goal of unlearning is to ensure an adversary with access to $\theta'$ cannot reconstruct information about $D^-$ via some readout function $f(\cdot)$. \\

\noindent \textbf{Definition 1.} Given an optimal unlearning algorithm $S_1(\theta, D)$, there is another function $S_2(\theta,  D^+)$ that does not depend on $D^-$ such that:
\begin{equation}
    KL(\mathcal{P}(f(S_1(\theta, D))||\mathcal{P}(f(S_2(\theta,D^+)))) = 0
\end{equation}
\label{KL-dif}
Where $KL$ stands for Kullback–Leibler divergence and $\mathcal{P(\cdot)}$ is the distribution of unlearned weights. Intuitively, the optimal unlearning outcome of $S_1(\theta, D)$ should be indistinguishable from a model that has never seen $D^-$. 
Satisfying this condition may be trivial, e.g. letting $S_1(\theta, D) = S_2(\theta, D^+)=c$ be constant. The challenge lies in how to do so while preserving as much knowledge as possible about $D^+$. We may define this relation from the perspective of Shannon Information \cite{golatkar2020eternal}:\\

\noindent \textbf{Proposition 1}
Let the forgetting set $D^-$ be a random variable such as a random sampling of $D$. Let $Y$ be an attribute of interest for $D^-$.

\begin{equation}
\begin{split}
    I(Y;f(S_1(\theta, D))) \leq \\
    \mathbb{E}_{D^-}[KL(\mathcal{P}(f(S_1(\theta,D ))||\mathcal{P}(f(S_2(\theta, D^+))))]
\end{split}
\end{equation}

Where $I(\cdot)$ represents mutual information.
In general, we may not know what reconstruction function an adversary will use, and hence this relation should be robust against every $f(\cdot)$. therefore, the following lemma is more practical:\\

\noindent \textbf{Lemma 1} For every $f(\cdot)$, we have:

\begin{equation}
    \begin{split}
        KL(\mathcal{P}(f(S_1(\theta,D ))||\mathcal{P}(f(S_2(\theta, D^+)))) \leq \\
        KL(\mathcal{P}(S_1(\theta,D )||\mathcal{P}(S_2(\theta, D^+)))
    \end{split}
\end{equation}

Thus, we should focus on minimizing the quantity.

$$KL(\mathcal{P}(S_1(\theta,D )||\mathcal{P}(S_2(\theta, D^+))) $$
which guarantees robustness to any readout function. Now we can show how to derive unlearning strategies using this idea.

\subsection{Unlearning for Convex Models}
\label{sec:noise-unlearn}
If the loss function is convex, e.g. quadratic, we can construct the following unlearning strategies:
$$S_1(\theta, D) = h(\mathcal{A}(D, \epsilon)) + n$$  $$S_2(\theta, D^+) = \mathcal{A}(D^+, \epsilon) + n'$$

where $\mathcal{A}(D, \epsilon)$ stands for a stochastic training algorithm with random seed $\epsilon$, 
$n, n' \sim \mathcal{N}(0, \Sigma)$ is Gaussian noise and $h(\cdot)$ is a deterministic function.
We have $\mathcal{P}(S_1(\theta,D ) \sim \mathcal{N}(h(\mathcal{A}(D, \epsilon)), \Sigma)$ and 
$\mathcal{P}(S_2(\theta, D^+)) \sim \mathcal{N}(\mathcal{A}(D^+, \epsilon),\Sigma)$. Then we have the following:

\begin{equation}
    KL(\mathcal{P}(S_1(\theta,D )||\mathcal{P}(S_2(\theta, D^+))) \leq 
    \frac{1}{2} \mathbb{E}_\epsilon[U^T\Sigma^{-1}U]   
\end{equation}

Where $U = h(\mathcal{A}(D, \epsilon))-\mathcal{A}(D^+, \epsilon)$.

This means that we can derive an upper bound for the complex KL quantity $KL(\mathcal{P}(S_1(\theta,D )||\mathcal{P}(S_2(\theta, D^+))$ by simply averaging the results of training and unlearning with different random seeds. In fact, this  KL quantity converges to zero if we simply apply the Newton update. Details can be found in Appendix \ref{proof_1}.

\subsubsection{Newton Update}
Without loss of generality, we aim to remove the last training sample $(x_n, y_n)$. Let the full dataset be denoted by $D$ and the left dataset $D^+ = D\setminus x_n$.
Firstly, the loss gradient with respect to $x_n$ can be calculated by $\Delta = \nabla \mathcal{L}(\theta x_n, y_n)$, where $\mathcal{L}$ is usually cross entropy loss and $\theta$ is the current model weight. Secondly, the Hessian of $\mathcal{L}(:, D^+)$ at $\theta$ is $H_{\theta} = \nabla^2 \mathcal{L}(\theta, D^+)$. Finally, the Newton update removal mechanism is as follows:

\begin{equation}
    S_1(\theta, D) = \theta + H_{\theta}^{-1} \Delta
\end{equation}

For shallow(convex) machine learning models, Newton update can perform unlearning with a bounded approximation error \cite{guo2019certified}.

\subsection{Gradient Ascent}
An intuitive way to perform unlearning is inverting the direction of gradients. Jang \textit{et al.}\cite{jang2022knowledge}
first studied this approach in the context of LLMs and we also include it due to its efficiency.
However, it has not yet been proven whether the gradient ascent is robust in maintaining accuracy on
$D^+$. 

As the opposite of the original training objective, gradient ascent (Algorithm \ref{algo:gradient-ascend}) aims to maximize the loss function such that the knowledge of certain samples is destroyed.

\begin{algorithm}[H]
    \caption{Gradient Ascent for LLM}
\begin{algorithmic}[1]
\Procedure{Gradient\_Ascent}{$D^-, \theta, l$}
\For{i = 1,2...n}
\State $\Delta \theta \leftarrow  L(B^-_{i}, \theta)$ \newline \hspace*{.2cm}\Comment{\textcolor{gray}{Gather the gradient for one batch in $D^-$}}
\State $\theta \leftarrow \theta  + l* \Delta \theta$   \Comment{\textcolor{gray}{Update parameters}}
\EndFor
\EndProcedure
\end{algorithmic}
\label{algo:gradient-ascend}    
\end{algorithm}

\section{Unlearning for LLMs}
However, LLMs
\cite{rashid2023fltrojan}  as well as other deep neural networks usually do not satisfy the convexity. To relax the assumption, we exploit the limited degree of freedom by
introducing noise into the general unlearning schema as discussed in Section \ref{sec:noise-unlearn}.

Consider noisy Newton update as follows:

\begin{equation}
   S_1(\theta, D) = \theta + H_{\theta}^{-1}\Delta + (\mu\sigma^2)^{\frac{1}{4}}H_{\theta}^{-\frac{1}{4}}
\end{equation}
\label{eq:noisy newton}

Here, $\mu$ controls the trade-off between residual information about the forgetting subset $D^-$, and accuracy on the remained data $D^+$. $
\delta$ reflects the error in approximating the SGD behavior with a continuous gradient flow. 

However, to calculate Hessian analytically requires $O(d^2)$ time, where $d$ is the size of the weight. For example, the embedding layer in GPT-2 small version has a size of $50257 \times 768 = 38597376$. Calculating the Hessian for this weight alone will require over $10^{15}$ operations. Not to mention that calculating the inverse of a matrix typically requires at least $O(d^{2.373})$ time \cite{Williams2014MultiplyingMI}.

\subsection{Inverse Empirical Fisher}
\label{sec:emp-fisher}
 Therefore, we estimate the inverse Hessian of LLMs via inverse empirical Fisher. Specifically, we employ the Woodbury/Sherman-Morrison (WSM) formula \cite{singh2020woodfisher}. Given a sum of $A + uv^T$ an invertible matrix $A$ and an outer product of vectors $u$ and $v$, the inverse $(A + uv^T)^{-1}$ can be exactly calculated as $A^{-1} - \frac{A^{-1}uv^T A^{-1}}{1 + v^TA^{-1}u}$. Combining the formula with empirical Fisher, we can obtain the following recursive formulation:

\begin{equation}
  F_{m}^{-1}(w) =   (F_{m-1}(w) + \frac{1}{m} \nabla \mathcal{L}(w, \hat{D})
  \nabla^T \mathcal{L}(w, \hat{D}))^{-1}
\end{equation}

Where $m$ is the number of gradients used for the approximation, equal to the
size of $\hat{D}$. With $F^{-1}_{0}(w) = \frac{1}{\lambda} I_d$, the above recursion can be rewritten as:
\begin{equation}
    F^{-1}_{m}(w) = \frac{1}{\lambda} I_d - \sum_{i=1}^{m} \frac{(F^{-1}_{i-1}(w) \nabla \mathcal{L}_{i}(w)) (F^{-1}_{i-1}(w) \nabla \mathcal{L}_{i}(w))^T }{m +  \nabla \mathcal{L}_{i}(w)^T F^{-1}_{i-1}  \nabla \mathcal{L}_{i}(w) }
\end{equation}

Here, $\lambda$ is an initialization parameter for $F_0^{-1}$.

\subsection{Efficient GPU Implementation}

It can even be prohibitively expensive to compute and store empirical Fisher $F^{-1}(w) \in \mathbb{R}^{d\times d}$ for LLMs due to the huge size of the model weights. However, by adapting the diagonal block-wise trick \cite{singh2020woodfisher}, empirical Fisher can be accurately 
approximated. Specifically, we can only focus on blocks of width $B$ along the main diagonal, which brings down the computation from quadratic $O(d^2)$ to linear $O(Bd)$.

On the implementation side, we have identified general hyper-parameters $B= 48$  for the block size, and $m= 1024$ for the maximum number of recursions which yield satisfying performances. A 16GB GPU can accommodate a 125M model (e.g. GPT2-small) with $B = 48$. As for large LLMs, we have designed a multi-GPU strategy to store Fisher matrices on different GPUs, which makes this approach scalable to state-of-the-art LLMs.

\subsection{Fisher Removal}
Although gradient descent is highly efficient, it does not account for information from $D^+$. Following the idea of noisy Newton update, we propose Fisher removal, which combines second-order information obtained from $D^+$ with first-order information obtained from $D^-$. 
 Formally, Fisher Removal is formulated as below:

$$S_1(\theta, D) = \theta + \gamma \hat{H}_{\theta}^{-1}\Delta 
$$

where unlearning rate $\gamma$ is adopted to adjust the unlearning effects for LLMs.

\begin{algorithm}[H]
\caption{Fisher Removal for LLM}
\begin{algorithmic}[1]

\Procedure{Fisher\_Removal}{$D^-, D^+, \theta, \lambda, \gamma$}

\For{i=1,2...,n} \Comment{\textcolor{gray}{Iterate through batches in $D^-$}}

\State $\Delta \theta \leftarrow  L(B^-_{i}, \theta)$ \newline \hspace*{.2cm} \Comment{\textcolor{gray}{Gather the gradient for one batch in $D^-$}}
 
\State $F_{0}^{-1} = \lambda^{-1} * I_d$ \newline \hspace*{.2cm}\Comment{\textcolor{gray}{Initialize Fisher Inverse with proper shape}}

\For{j=1,2...m} 
\newline \hspace*{.2cm}\Comment{\textcolor{gray}{Iterate through batches in $D^+$}}

   \State $\Delta_{B^+_j} \leftarrow  L(B^+_j, \theta)$  \newline \hspace*{.1cm}\Comment{\textcolor{gray}{Gather gradient for one batch in $D^+$}}
    \State $F_{j}^{-1} \leftarrow update(F_{j-1}^{-1}, \Delta_{B^+_j} )$ \newline \hspace*{.2cm} \Comment{ \textcolor{gray}{Invoke Empirical Fisher recursion}}
\EndFor

\State $\hat{H}^{-1} = F_{m}^{-1}$
\State $\theta \leftarrow \theta + \gamma \hat{H}^{-1} \Delta \theta$ \Comment{\textcolor{gray}{Update parameters}}
\EndFor
\EndProcedure
\end{algorithmic}
\label{algo:fisher-removal}
\end{algorithm}

\subsection{Fisher Forgetting}
\label{sec:fisher-forgetting}
Since $\hat{H}^{-1}$ is an empirical approximation, it introduces additional uncertainty into the unlearning process, which can cause an unexpected loss of accuracy on $D^+$. Therefore, we present Fisher Forgetting to perturb the memory of $D^-$ by adding Gaussian noise to the neurons \cite{golatkar2020eternal}.

$$S_1(\theta, D) = \theta + (\mu\sigma^2)^{\frac{1}{4}}H_{\theta}^{-\frac{1}{4}} \odot M $$

Where $M \sim \mathcal{N}(0,1)$ is Gaussian noise, $\mu , \sigma$ are parameters introduced in Section \ref{eq:noisy newton}.

\begin{algorithm}[H]
\caption{Fisher Forgetting for LLM}
\begin{algorithmic}[1]

\Procedure{Fisher\_Forgetting}{$D^-, D^+, \theta, \lambda, \mu, \sigma$} 

\For{i=1,2...,n} \Comment{\textcolor{gray}{Iterate through batches in $D^-$}}

\State $\Delta \theta \leftarrow  L(B^-_{i}, \theta)$\newline \hspace*{.2cm} \Comment{\textcolor{gray}{Gather the gradient for one batch in $D^-$}}
 
\State $F_{0}^{-1} = \lambda^{-1} * I_d$\newline \hspace*{.2cm} \Comment{  \textcolor{gray}{Initialize Fisher Inverse with proper shape}}

\For{j=1,2...m} \newline \hspace*{.2cm}\Comment{\textcolor{gray}{Iterate through batches in $D^+$}}
    \State $\Delta_{B^+_j} \leftarrow - L(B^+_j, \theta)$  \newline \hspace*{.2cm}\Comment{\textcolor{gray}{Gather gradient for one batch in $D^+$}}
    \State $F_{j}^{-1} \leftarrow update(F_{j-1}^{-1}, \Delta_{B^+_j} )$  \newline \hspace*{.2cm}\Comment{\textcolor{gray}{Invoke Empirical Fisher recursion}}
\EndFor

\State $\hat{H}^{-1} = F_{m}^{-1}$
\State $M \sim \mathcal{N}(0, 1) $ \hspace*{.2cm} \Comment{\textcolor{gray}{Sampling Gaussian Noise}}
\State $\theta \leftarrow \theta + (\mu \sigma^2)^{\frac{1}{4}}\hat{H}^{- \frac{1}{4}} \odot M $\newline \hspace*{.2cm} \Comment{\textcolor{gray}{Update parameters}}
\EndFor
\EndProcedure
\end{algorithmic}
\label{algo:fisher-forgetting}
\end{algorithm}

\noindent \textbf{Property of Fisher Forgetting}

Assuming $||H_{\theta}^{-\frac{1}{4}} ||_2 \leq 1$, which can be easily satisfied if $\lambda$ is set to $1$ when initializing $F_0^{-1}$, we can derive the following property:

$$||(\mu\sigma^2)^{\frac{1}{4}}H_{\theta}^{-\frac{1}{4}} \odot M ||_2 \leq (\mu\sigma^2)^{\frac{1}{4}} ||M||_2$$

Considering Fisher Forgetting has been performed $k$ times in a roll on a LLM parameterized by $\theta$, 
the sum of all $k$ updates to the weights can be calculated by:

$$\sum_{i=1}^{k} (\mu\sigma^2)^{\frac{1}{4}}H_{\theta}^{-\frac{1}{4}} \odot M_i = \sum_{i=1}^{k} \mathcal{N}(0, ((\mu\sigma^2)^{\frac{1}{4}}H_{\theta}^{-\frac{1}{4}}))^2) $$

Here $M_i$ are i.i.d Gaussian variables $\sim \mathcal{N}(0,1)$.
By the property of Gaussian distributions:
$$ \sum_{i=1}^{k} \mathcal{N}(0, ((\mu\sigma^2)^{\frac{1}{4}}H_{\theta}^{-\frac{1}{4}}))^2) \sim N(0,\sum_{i=1}^{k} ((\mu\sigma^2)^{\frac{1}{4}}H_{\theta}^{-\frac{1}{4}}))^2)$$

Again, with $||H_{\theta}^{-\frac{1}{4}} ||_2 \leq 1$, the quantity can be approximated as $\mathcal{N}(0, k(\mu\sigma^2)^{\frac{1}{2}})$. This means the cumulative updates of $k$ consecutive Fisher Forgetting can be bounded by a new Gaussian distribution $\mathcal{N}(0, k(\mu\sigma^2)^{\frac{1}{2}})$, where $\mu, \sigma << 1$.
As a result, the model performance can be largely preserved.

\section{Experimental Setup}
\begin{figure}[h]
\includegraphics[width=8cm]{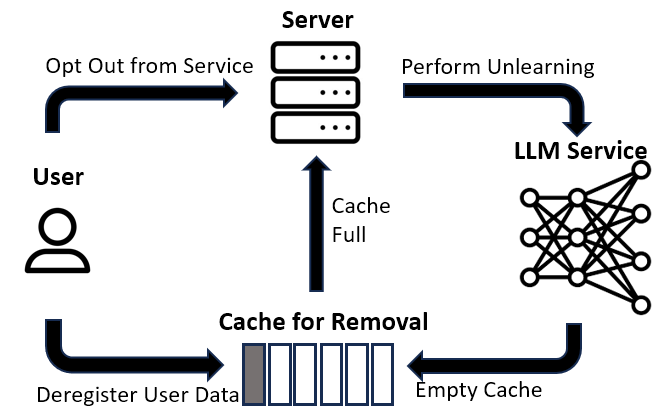}
\caption{The cycle of unlearning of the LLM service. The model has been trained on the user database for downstream tasks. Since user requests arrive stochastically, a cache is used to store them. Once the cache is full or the waiting time has exceeded the threshold, the server will invoke the unlearning process and clean the cache to free up space. 
Without the loss of generality, we assume the size of the cache is 128 in our experiments.
}
\label{fig:Unlearning-illus}
\end{figure}

Figure \ref{fig:Unlearning-illus} illustrates how a service provider handles incoming removal requests in our setting. In the life cycle of an LLM service, such requests for removal are likely to arrive stochastically. To protect users' right to be forgotten, the server may need to perform unlearning more than once, depending on the predetermined terms.

In our experiments, we consider two scenarios of unlearning: 1) perform only one unlearning cycle (128 samples), 2) perform two unlearning cycles (256 samples) consecutively. 

\subsection{Datasets}
We evaluate LLMs on four general NLP datasets to show the trade-offs between unlearning effects and resulting performances on downstream tasks. 

\subsubsection{ARC-Easy.}
This dataset contains $5000$ in grade-school level, multiple-choice scientific questions. For example, given the question of  ``Which piece of safety equipment is used to keep mold spores from entering the respiratory system?'' and options ``[A: safety goggles, B: breathing mask, C: rubber gloves, D: lead apron]'', the answer is B.
The dataset is split into train/val/test sets with $2251$, $570$, and $2376$ questions respectively.

\subsubsection{Piqa.}
It is a binary physical question-answering dataset. To solve the questions correctly, the models are required to understand physical commonsense. An example question is ``To separate egg whites from the yolk using a water bottle, you should ...'', and the answer is ``Squeeze the water bottle and press it against the yolk. Release, which creates suction and lifts the yolk. "
This dataset consists of $16113$ and $1838$ questions for training and validation respectively.

\subsubsection{MathQA.}
It is a large-scale dataset of math world problems, which has provided the questions, options, answers as well as rationales. The train/val/test splits have $29837$, $4475$, and $2985$ questions.

\subsubsection{Lambada.}
This dataset evaluates the language modeling capacity by the task of last-word prediction. Human subjects can guess the last word with at least 50 tokens of context, which requires long-term dependencies. The dataset contains 10022 passages from BookCorpus and is further divided into 4869 for development and 5153 for testing.

\subsection{Evaluation Metrics}
We propose to evaluate unlearning algorithms from three dimensions: 1) the removal of selected data, 2) the preservation of model utility, and 3) the used time as in \cite{jang2022knowledge}.
For simplicity, let $\theta$ be the parameters of an initial language model requiring unlearning, 
$D^-$ be a batch of data to unlearn,
$D_{test}$ be a standalone test dataset, $U$ be an unlearning algorithm, and $\theta'$ be the model after being unlearned.

\subsubsection{Efficacy.}
The most important property of an unlearning algorithm is ensuring successful removal. Although certified unlearning \cite{guo2019certified} can guarantee this removal, there is no such guarantee for non-convex models. Perplexity or validation loss may not be sufficient to distinguish $\theta$ and $\theta'$ \cite{mireshghallah2022quantifying}. Therefore, we refer to \textit{exposure metrics} \cite{carlini2019secret} to measure the efficacy of unlearning quantitatively.

\begin{equation}
    exp_{\theta}(s) = \log_2 (|Q|) - \log_2rank_{\theta}(s)
\end{equation}

Here, $s$ is a text sample and $Q$ is a set of possible sequences with the same length. The $rank_{\theta}(s)$ function returns the rank of $s$ given model $\theta$ and the set $Q$.

Based on the exposure metrics of a single sample, we propose $\Delta_{exp}(\theta, \theta')$ to measure the change of expected exposure values after unlearning is performed. 

\begin{equation}
  \Delta_{exp}(\theta, \theta') =  \mathbb{E}_{s\in D^-} [exp_{\theta'}(s) - exp_{\theta}(s)] 
\end{equation}

The above metrics reflect the extent to which the unlearning algorithm $U$ erases the set $D^-$.\\

\noindent \textbf{Approximating Exposure:}
Given the length $l$, it is not practical to iterate through all combinations of tokens in the model vocabulary $V$, which leads to $|V|^l$ possibilities. We choose to approximate the exposure by sampling \cite{carlini2019secret}. Specifically, we refer to a standalone corpus and sample a set $S$ of sequences with length $l$.  
$S$ can be viewed as a random sampling of the full space $Q$ such that $|S|<<|Q|$.

\subsubsection{Fidelity.}
Although the removal of $D^-$ matters, the unlearning algorithm $U$ should also largely protect the performance of the model $\theta'$. Without loss of generality, we use accuracy to evaluate model performance. Formally, let
$acc_{\theta}(D_{test})$  stand for the accuracy of model $\theta$ on test set $D_{test}$,  $acc_{\theta'}(D_{test})$ is expected to be close to the initial model $\theta$.
Similarly, we propose $\Delta_{acc}(\theta, \theta')$ to reflect the change of accuracy.
\begin{equation}
    \Delta_{acc}(\theta, \theta') = acc_{\theta'}(D_{test}) - acc_{\theta}(D_{test}) 
\end{equation}

\subsubsection{Efficiency.}
Finally, the unlearning algorithm should be efficient compared with retraining. As a result, we show both the time complexity and actual running time of each unlearning method in our experiments.

\subsection{Unlearning Settings}

\noindent \textbf{Target LLMs.}
We test the unlearning methods on both GPT-Neo \cite{black2022gpt} and OPT LMs \cite{zhang2022opt} as in \cite{jang2022knowledge}. For each model, three variants (125M, 1.3B, 2.7B) are included.

\begin{table}[!htb]
    \caption{The Comparison of Selected Existing Unlearning Methods}
    \resizebox{.47\textwidth}{!}{ 
    
    \begin{tabular}{|l|c|c|c|}
    \hline
     Method  &  Data  & Architecture & Reproducible/ \\
     &  Engineering & Engineering & Open-Sourced\\
     \hline
     Retraining  & x & x & $\checkmark$ \\
     Finetuning & x& x & $\checkmark$ \\
     Gradient Ascent \cite{jang2022knowledge} & x & x & $\checkmark$ \\
     SISA \cite{bourtoule2021machine}& $\checkmark$ & x & $\checkmark$\\
     Revise \& Finetuning \cite{eldan2023s}& $\checkmark$ & x & x \\
     EUL \cite{chen-yang-2023-unlearn} & x & $\checkmark$ & x\\
     Ours* & x & x & $\checkmark$\\
    \hline
    \end{tabular}
    }
    
    \label{tab:compare_unlearn}
\end{table}

\noindent \textbf{Unlearning Baselines.}
We show the comparison of selected unlearning methods in 
Table \ref{tab:compare_unlearn}.
Only methods that can be reproducible on any dataset or LLM are considered in our experiments. However, we exclude SISA \cite{bourtoule2021machine} because dividing the training data into disjoint shards and training a LLM on each shard will compromise the performance of LLMs, especially when the number of shards is large. 
As a result, in addition to the aforementioned unlearning Algorithm \ref{algo:gradient-ascend}-\ref{algo:fisher-forgetting}, we also consider retrain and finetuning as unlearning baselines:
\begin{itemize}
    \item \textbf{Retraining:} We discard the current model and retrain the LLM using $D^+$ for 5 epochs.
    It is the benchmark for our evaluation of other baselines.
    \item \textbf{Finetuning:} We continue to finetune the current model on $D^+$ for 1 epoch. Intuitively, the model's memory of $D^-$ will be weakened.
\end{itemize}

\subsection{Hyperparameters}
\begin{itemize}
    \item For all the LLMs in our experiments, we set the coefficient for gradient ascent $l$ as $5e-5$, which is the same as the learning rate. 
As for second-order methods, $\gamma$ is set as $2.5e-4$. $\mu$ and $\sigma$ are both set to $1e-3$. 
\item For empirical Fisher, $\lambda $ and $m$ are set to $1$ and $1024$, respectively.

\item To implement exposure metrics, we refer to the European Court of Human Rights (ECHR) corpus\footnote{https://archive.org/details/ECHR-ACL2019}. The size of subspace $S$ is set to $1000$. 
\end{itemize}

Note that $\gamma$, $\mu$ and $\sigma$ are chosen through grid search where $\gamma, \mu, \sigma \in [1e-4, 1e-3]$.

\section{Experimental Results}
In this section, we evaluate all unlearning baselines on four datasets respectively and further highlight the overall performances of each baseline on efficacy, fidelity, and efficiency.

\subsection{Results \&  Discussions}
To improve the interpretability of the results, we utilize color coding\footnote{Blue cells stand for that corresponding methods are superior to retraining in the current settings while gray cells mark catastrophic unlearning outcomes, i.e. the accuracy drops significantly ($> 10\%$) or the exposure metric increases instead of decreasing ($\Delta_{EXP}>0$).} to emphasize on distinctive values in Table 
\ref{tab:Unlearning_ARC}, \ref{tab:Unlearning_Piqa},
\ref{tab:Unlearning_MathQA}.
Due to space constraints, the results of the Lambada dataset (Table \ref{tab:Unlearning_Lambada}) can be found in the Appendix.

\begin{table*}[t]
    \centering
    \caption{Unlearning Results on ARC Dataset}
\resizebox{.8\textwidth}{!}{ 
 \begin{tabular}{cc|cccc|cccc}
    \hline
    \textbf{Parameter}  & \textbf{Unlearning} &\multicolumn{4}{c|}{\textbf{128 samples}} & \multicolumn{4}{c}{\textbf{256 samples}} 
      \\ \cline{3-10}
      & & \multicolumn{2}{c}{NEO} & \multicolumn{2}{c|}{OPT}&
      \multicolumn{2}{c}{NEO} & \multicolumn{2}{c}{OPT}\\
      \hline
        & &  $\Delta_{ACC}$   & $\Delta_{EXP}$  & $\Delta_{ACC}$  & $\Delta_{EXP}$ &  $\Delta_{ACC}$   & $\Delta_{EXP}$  & $\Delta_{ACC}$  & $\Delta_{EXP}$  \\ \hline

       \multirow{6}{*}{125M} 
       &  Retraining & -0.00 & -0.24 &-0.71 &-0.05 &-0.30 &-0.21 & +0.04&-0.10 \\
       & Finetuning & \cellcolor{blue!10}+2.00 & \cellcolor{gray!25}+0.04 & \cellcolor{blue!10}+0.25&-0.00 &\cellcolor{blue!10}+1.20 & -0.00 
       & -0.08 & -0.08 \\
      
      & Gradient Ascent & -0.74& -0.12 &  -6.35&\cellcolor{blue!10}-0.40 & \cellcolor{gray!25} -10.0& \cellcolor{blue!10}-1.24 & 
      \cellcolor{gray!25}-12.20 &\cellcolor{blue!10} -6.08\\
      & Fisher Removal &-2.20 &\cellcolor{blue!10} -0.37 &-4.84 & \cellcolor{blue!10}-0.07&-4.40 & \cellcolor{blue!10}-1.24  
      &\cellcolor{gray!25}-10.27 & \cellcolor{blue!10}-5.93\\
      & Fisher Forgetting &-0.60 & -0.04 &\cellcolor{blue!10}-0.67 &-0.01 &-1.40 &-0.11 & -1.18 & -0.08\\
        \hline
      \multirow{6}{*}{1.3B} 
      &  Retraining & -0.80 & -0.48 & +0.17&-0.05 &+0.51 &-0.37 & -1.48 &-0.06 \\
      & Finetuning & \cellcolor{blue!10} -0.37 & \cellcolor{gray!25} +0.02 & \cellcolor{blue!10} +1.10 &
      -0.01 
      &\cellcolor{blue!10}+1.48 & \cellcolor{gray!25}+0.02
      &\cellcolor{blue!10}-0.42 & -0.00\\

      & Gradient Ascent & \cellcolor{blue!10}-0.04 & -0.01&-0.80 &-0.02 & -0.00& -0.02 & \cellcolor{gray!25}-20.12 & \cellcolor{blue!10}-1.22\\
      & Fisher Removal &\cellcolor{blue!10}-0.21 &-0.04 &-7.93 &\cellcolor{blue!10}-0.20 & -2.52&-0.23 & \cellcolor{gray!25}-10.79& \cellcolor{blue!10}-4.09 \\
      & Fisher Forgetting & \cellcolor{blue!10}-0.71& -0.02&-0.34 & -0.02
      &-0.96 & -0.02  & -2.52 & -0.03\\
        
      \hline
      
      \multirow{6}{*}{2.7B} 
       &  Retraining & -1.01 & -0.53&
       +0.59 &-0.10 & +0.80&-0.47&+0.21 & -0.17\\
      & Finetuning &\cellcolor{blue!10}-0.51 &-0.08 &\cellcolor{blue!10}+1.14
      &-0.02 & \cellcolor{blue!10}+1.22 &-0.05 & \cellcolor{blue!10}+2.28 & -0.03\\
     
      & Gradient Ascent &-1.52 & -0.14&\cellcolor{gray!25}-21.46
       &\cellcolor{blue!10} -6.28&-1.10&-0.28&\cellcolor{gray!25}-21.42 &\cellcolor{blue!10}-6.14\\
      & Fisher Removal  &-4.04 &-0.27 &-4.68
      &\cellcolor{blue!10} -1.40&\cellcolor{gray!25} -14.33&\cellcolor{blue!10}-3.40 & \cellcolor{gray!25}-20.38 & \cellcolor{blue!10}-6.03\\
      & Fisher Forgetting &-1.52 &-0.08 & -0.33&-0.03 &-2.23&-0.10&-1.60 & -0.03\\

      \hline  
    \end{tabular} 
   }
    
    \label{tab:Unlearning_ARC}
\end{table*}

\begin{table*}[t]
    \centering
    \caption{Unlearning Results on Piqa Dataset}
\resizebox{.8\textwidth}{!}{ 
 \begin{tabular}{cc|cccc|cccc}
    \hline
    \textbf{Parameter}  & \textbf{Unlearning} &\multicolumn{4}{c|}{\textbf{128 samples}} & \multicolumn{4}{c}{\textbf{256 samples}} 
      \\ \cline{3-10}
      & & \multicolumn{2}{c}{NEO} & \multicolumn{2}{c|}{OPT}&
      \multicolumn{2}{c}{NEO} & \multicolumn{2}{c}{OPT}\\
      \hline
        & &  $\Delta_{ACC}$   & $\Delta_{EXP}$  & $\Delta_{ACC}$  & $\Delta_{EXP}$ &  $\Delta_{ACC}$   & $\Delta_{EXP}$  & $\Delta_{ACC}$  & $\Delta_{EXP}$  \\ \hline

       \multirow{6}{*}{125M} 
        &  Retraining & -0.50 & -0.03  &+1.14 &-0.02 & -0.06 & -0.14 & +1.09&-0.01\\
       & Finetuning & \cellcolor{blue!10}-0.20 &\cellcolor{gray!25}+0.17  & +0.27& \cellcolor{gray!25}+0.07 &\cellcolor{blue!10}+1.20 & \cellcolor{gray!25} +1.14 &
       \cellcolor{blue!10} +1.30 & \cellcolor{gray!25}+0.06\\

      & Gradient Ascent & -3.50 &\cellcolor{blue!10}-3.03 & \cellcolor{gray!25}-10.29 &\cellcolor{blue!10} -3.32 &  \cellcolor{gray!25} -12.9&\cellcolor{blue!10} -3.53
      &\cellcolor{gray!25}-10.07 & \cellcolor{blue!10}-1.12\\
      & Fisher Removal &\cellcolor{blue!10}-0.36 & \cellcolor{blue!10}-1.95 & -5.62& \cellcolor{blue!10} -4.84 & -2.65& \cellcolor{blue!10}-4.57 
      &-9.67 & \cellcolor{blue!10} -4.28\\
      & Fisher Forgetting & \cellcolor{blue!10}+0.07& \cellcolor{blue!10}-0.06
       &-0.17 &\cellcolor{blue!10} -0.02 &-0.85 &-0.05&
       -1.47 & \cellcolor{blue!10}-0.02\\
        \hline
      \multirow{6}{*}{1.3B} 
      &  Retraining & +0.66& -0.10 & -1.52&-0.05 &+1.15 &-0.15
      & -0.76 &-0.00 \\
      & Finetuning &\cellcolor{blue!10}+1.15 & \cellcolor{gray!25} +0.13 & \cellcolor{blue!10}-0.70&\cellcolor{gray!25}+0.05 & \cellcolor{blue!10}+1.47& \cellcolor{gray!25}+0.09 
      & -1.14 & \cellcolor{gray!25}+0.03\\

      & Gradient Ascent &-0.16 &\cellcolor{blue!10}-0.14 & \cellcolor{blue!10}-1.03& -0.03&  \cellcolor{gray!25}-10.35&\cellcolor{blue!10}-4.86
      & \cellcolor{gray!25}-17.73 & \cellcolor{blue!10}-4.29\\
      & Fisher Removal &+0.22 &\cellcolor{blue!10}-0.16 & -5.13&\cellcolor{blue!10}-0.84 &-2.28 & \cellcolor{blue!10}-3.39 
      & -7.95 & \cellcolor{blue!10}-5.91\\
      & Fisher Forgetting & -0.22& -0.03&\cellcolor{blue!10}-0.41 &-0.01 &-0.27&-0.01 & \cellcolor{blue!10}-0.68& \cellcolor{blue!10}-0.01\\
        
      \hline
      
      \multirow{6}{*}{2.7B} 
      &  Retraining &-0.33&-0.15 &-0.49 & -0.07
      &-0.76 &-0.22& -0.54&-0.02\\
      & Finetuning & -0.44& \cellcolor{gray!25}+0.01 & -1.19&\cellcolor{gray!25}+0.12 &-1.19
      & \cellcolor{gray!25}+0.12&
      -0.76& \cellcolor{gray!25}+0.15 \\
      & Gradient Ascent&\cellcolor{blue!10} -0.33 & \cellcolor{blue!10}-0.28 & -5.11&\cellcolor{blue!10}-0.27 & -6.97&  \cellcolor{blue!10}-5.09
      &\cellcolor{gray!25}-18.49 &\cellcolor{blue!10}-5.57\\
      & Fisher Removal & -1.36 &\cellcolor{blue!10}-0.79 & -6.97&\cellcolor{blue!10}-1.12
       & -6.08& \cellcolor{blue!10}-4.77
       &  \cellcolor{gray!25}-12.60&\cellcolor{blue!10}-3.40\\
      & Fisher Forgetting& -0.60&-0.00 &\cellcolor{blue!10}-0.45 &-0.01
      & \cellcolor{blue!10} -0.63 & -0.09&-1.14&\cellcolor{blue!10}-0.02\\

      \hline  
    \end{tabular} 
   }
    
    \label{tab:Unlearning_Piqa}
\end{table*}

\begin{table*}[t]
    \centering
    \caption{Unlearning Results on MathQA Dataset}
\resizebox{.8\textwidth}{!}{ 
 \begin{tabular}{cc|cccc|cccc}
    \hline
    \textbf{Parameter}  & \textbf{Unlearning} &\multicolumn{4}{c|}{\textbf{128 samples}} & \multicolumn{4}{c}{\textbf{256 samples}} 
      \\ \cline{3-10}
      & & \multicolumn{2}{c}{NEO} & \multicolumn{2}{c|}{OPT}&
      \multicolumn{2}{c}{NEO} & \multicolumn{2}{c}{OPT}\\
      \hline
        & &  $\Delta_{ACC}$   & $\Delta_{EXP}$  & $\Delta_{ACC}$  & $\Delta_{EXP}$ &  $\Delta_{ACC}$   & $\Delta_{EXP}$  & $\Delta_{ACC}$  & $\Delta_{EXP}$  \\ \hline

       \multirow{6}{*}{125M} 
       &  Retraining & +0.40 & -0.03 &-0.51 &-0.01 &+0.20 & -0.02 &-0.04&-0.00\\
       & Finetuning & -0.40 &\cellcolor{gray!25} +0.31 & -0.61&\cellcolor{gray!25}+0.01 &-0.30 & \cellcolor{gray!25}+0.28 
       &-0.27&\cellcolor{gray!25}+0.02\\

      & Gradient Ascent & -1.90 & \cellcolor{blue!10}-0.23 &\cellcolor{blue!10}+0.59 &\cellcolor{blue!10}-6.49 & -2.50& \cellcolor{blue!10}-2.43
      &\cellcolor{blue!10} +1.04 & 
      \cellcolor{blue!10} -6.30\\
      & Fisher Removal &-0.56 & \cellcolor{blue!10}-1.13 &\cellcolor{blue!10}+0.63 &\cellcolor{blue!10}-5.88 & -1.33& \cellcolor{blue!10}-4.63
      &\cellcolor{blue!10}  +0.80 
      &\cellcolor{blue!10} -3.31\\
      & Fisher Forgetting &+0.11 &\cellcolor{blue!10}-0.03 & \cellcolor{blue!10}+0.83&\cellcolor{blue!10}-0.02 &\cellcolor{blue!10}+1.38 &\cellcolor{blue!10}-0.02
      & \cellcolor{blue!10} +0.03
      &-0.00\\
        \hline
      \multirow{6}{*}{1.3B} 
      &  Retraining & +1.94& -0.02& +0.47&-0.00 & +1.07&-0.00
      &+0.64 & -0.00\\
      & Finetuning & +1.25 & \cellcolor{gray!25} +0.32 & +0.07&\cellcolor{gray!25}+0.03 &
      \cellcolor{blue!10} +1.14 &
      \cellcolor{gray!25}+0.19
      & +0.10 & \cellcolor{gray!25}+0.03\\

      & Gradient Ascent &-1.31 &\cellcolor{blue!10}-0.11 & -1.34&\cellcolor{blue!10} -3.05&-2.48 & \cellcolor{blue!10} -3.87
      &-1.26& \cellcolor{blue!10}-1.28\\
      & Fisher Removal  &-0.30 &\cellcolor{blue!10}-0.12 & -0.36&\cellcolor{blue!10}-2.11 &-3.69&\cellcolor{blue!10} -4.42 & -1.07 & \cellcolor{blue!10}-4.39 \\
      & Fisher Forgetting  &-0.37 &\cellcolor{blue!10}-0.02 & -0.10&\cellcolor{blue!10}-0.01 &-1.04&-0.00
      &-0.43 & \cellcolor{blue!10}-0.02\\
        
      \hline
      
      \multirow{6}{*}{2.7B} 
      &  Retraining &-0.77 &-0.04 &+0.10 &-0.03 
      & -0.44 &-0.11 &+0.37 &-0.02\\
      & Finetuning &\cellcolor{blue!10} +1.13 & \cellcolor{gray!25}+0.33 &
     \cellcolor{blue!10}+0.20 &\cellcolor{gray!25}+0.07 & \cellcolor{blue!10} +1.13&
      \cellcolor{gray!25}+0.22
      &\cellcolor{blue!10} +0.84&\cellcolor{gray!25}+0.08\\
      & Gradient Ascent &-1.58 & \cellcolor{blue!10}-0.08  & +0.07&\cellcolor{blue!10}-0.03
      & -3.46& \cellcolor{blue!10} -1.79
      & -0.43&\cellcolor{blue!10}-1.63\\
      & Fisher Removal  &-3.49 & \cellcolor{blue!10}-0.71 & -0.50&\cellcolor{blue!10}-0.32
      &-5.64 &\cellcolor{blue!10} -5.21
      & \cellcolor{blue!10}+1.51&\cellcolor{blue!10}-6.28\\
      & Fisher Forgetting &\cellcolor{blue!10}+0.23 &-0.00& -0.16&\cellcolor{gray!25}+0.01
      & -0.67&-0.01
      &-0.23 &-0.01\\

      \hline  
    \end{tabular} 
  }
\label{tab:Unlearning_MathQA}
\end{table*}

\subsubsection{Efficacy Evaluation.}
With retraining as the benchmark,
we observe that Fisher Removal can effectively reduce exposure values. For example, in Table \ref{tab:Unlearning_Piqa}, 
Fisher Removal achieves a stronger efficacy guarantee than retraining in all the scenarios, as indicated by blue cells. Fisher Forgetting is less aggressive in terms of erasing the samples, which is consistent with its design.

However, gradient ascent cannot
provide an efficacy guarantee as robust as Fisher Removal. In Table \ref{tab:Unlearning_ARC},  the removal effects of gradient ascent are apparently  
insufficient for GPT-NEO models compared with Fisher Removal.
Meanwhile, finetuning frequently causes the samples that should be removed to be more exposed to the adversary, especially in Table 
\ref{tab:Unlearning_MathQA}.
It shows that finetuning, which is an intuitively plausible unlearning approach, actually does not erase the information from the samples steadily.

\subsubsection{Fidelity Evaluation}

In the case of a single unlearning cycle (128 samples), finetuning and Fisher Forgetting are relatively closer to retraining in accuracy. 
Fisher Removal causes slightly more degradation due to a more aggressive parameter update strategy. However, on closer inspection, gradient ascent causes multiple catastrophic outcomes as shown in Table \ref{tab:Unlearning_ARC} \& \ref{tab:Unlearning_Piqa}, suggesting gradient ascent is the least robust regarding retaining the model fidelity.

As for the case of two unlearning cycles (256 samples), finetune and Fisher Forgetting still maintain the accuracy as well as retraining. While Fisher Removal is relatively less utility-preserving compared with Fisher Forgetting. As discussed in Section \ref{sec:fisher-forgetting}, Fisher Forgetting has the property of supporting more unlearning cycles. Therefore, We further extend the cycles of unlearning and report the fidelity in Section \ref{extended-unlearning}.

\subsubsection{Efficiency Evaluation.}

We first formally present the time complexity of each method. Let the size of $D^-$ be $t$ and the size of $D^+$ be $r$  respectively. Besides, recall that $m$ is the number of recursions to estimate second-order information in Section \ref{sec:emp-fisher}. 

For simplicity, we assume the time cost for one backpropagation pass is $O(1)$. The time complexity of each method is shown in Table \ref{tab:runtime analysis}. It is noticeable that gradient ascent is the most efficient method among them given $t<<r$ can be easily satisfied. Moreover, the complexity of second-order methods is $O(tm)$. In fact, if the dataset is small such that $tm > r$, second-order methods are likely to be expensive. In contrast, when the dataset is large (which is more realistic for LLM applications),  second-order methods can be relatively more efficient.\\

\begin{table}[t]
\caption{Time complexity of each unlearning baseline}
    \centering
    \begin{tabular}{|c|c|c|}
    \hline
     Method   &  Time Complexity  \\ \hline
     Retraining  & $O(r)$ \\\hline
      Finetuning   & $O(r)$\\\hline
      Gradient Ascent & $O(t)$\\\hline
      Fisher Removal & $O(tm)$\\\hline
      Fisher Forgetting & $O(tm)$ \\
      \hline
    \end{tabular}
    
    \label{tab:runtime analysis}
\end{table}

\begin{figure}[t]
\includegraphics[width=8.5cm]{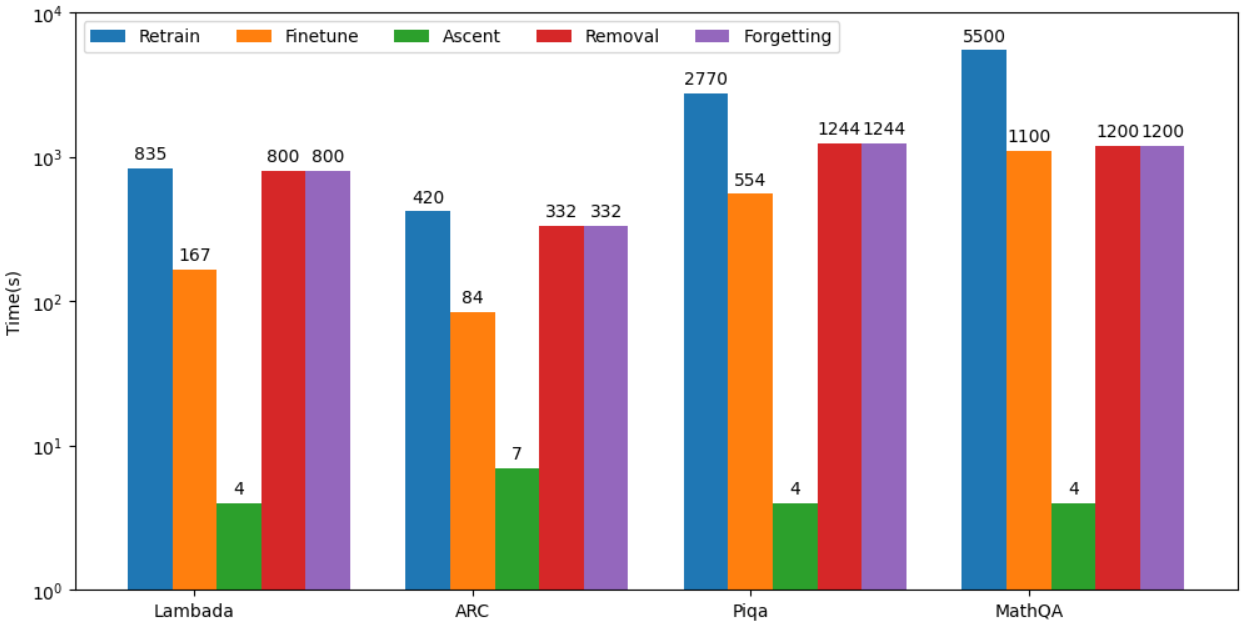}
\caption{The runtime of each unlearning method on GPT-NEO-125M.}
\label{fig:runtime}
\end{figure}

\noindent \textbf{Runtime Comparison.}
In addition to the time complexity, we also show the runtime of GPT-NEO-125M on various datasets in Figure \ref{fig:runtime}. Note that ARC ($\sim 2k$) and Lambada ($\sim 5k$) datasets are small, while Piqa ($\sim 16k$) and
MathQA ($\sim 30k$) datasets are relatively larger.

In general, the results are consistent with previous analysis.
Although second-order methods are almost as expensive as retraining on ARC and Lambada datasets, their runtime is close to finetuning on the MathQA dataset.

\subsubsection{Overall Evaluation.}
To understand the performances more comprehensively,
we aggregate the results on individual datasets by: 1)  calculating the rank of each method in each scenario\footnote{The range is within 1 to 5, with the best method ranked $1_{st}$ }, and 2) reporting the mean and standard deviation of the ranks for each method. 
Their overall rankings on different dimensions are shown in Table \ref{tab:rank}. Note that their performances on the two target LLMs are similar. Therefore, we conclude the behaviors of each unlearning approach as follows:
\begin{itemize}
    \item \textbf{Retraining} achieves the optimal balance between efficacy and fidelity as the benchmark for unlearning. However, it is the least efficient baseline as expected. 
    \item \textbf{Finetuning} cannot provide any efficacy guarantee, therefore it is not suitable for the purpose of unlearning.
    \item \textbf{Gradient Ascent} is unstable with high variations in both efficacy and fidelity. For example, its fidelity rank on the GPT-NEO models yields a standard deviation of 2.69.
    However, it can still serve as a baseline due to its extreme efficiency.
    
    \item \textbf{Fisher Removal}  demonstrates the strongest efficacy by outperforming both retraining and gradient ascent. However, its fidelity score is compromised because of the aggressive removal effect.
    
    \item \textbf{Fisher Forgetting} is on contrary to fisher removal. It reduces the strength of erase in exchange for more robust accuracy on the main task.
\end{itemize}

\begin{table*}[t]
\centering
\caption{Overall Rankings of all Unlearning Methods}
\resizebox{.8\textwidth}{!}{ 
\begin{tabular}{|c|ccc|ccc|}
\hline
 & \multicolumn{3}{c|}{NEO} & \multicolumn{3}{c|}{OPT} \\
 \hline
   Methods  & Efficacy($\downarrow$) & Fidelity($\downarrow$)  &  Efficiency($\downarrow$) &Efficacy($\downarrow$) & Fidelity($\downarrow$)  &  Efficiency($\downarrow$) \\
   \hline
   \hline
   Retraining  & 2.54+-0.91 &2.21+-1.08& 5+-0.00  &  2.92+-0.86&2.04+-1.02  & 5+-0.00\\
   Finetuning &4.75+-0.52  & 2.00+-0.96& 2+-0.00 & 4.88+-0.33& 2.00+-1.09 & 2+-0.00\\
   Gradient Ascent & 2.29+-1.06& 4.17+-2.69&1+-0.00 & 1.88+-1.05&4.42+-1.00 & 1+-0.00\\
   Fisher Removal &1.33+-0.47  &4.08+-1.08&3+-0.00 & 1.58+-0.64& 3.92+-0.95& 3+-0.00\\
   Fisher Forgetting &3.71+-0.61 &2.79+-1.12 & 3+-0.00 & 3.58+-0.49&2.63+-0.86 & 3+-0.00\\
   \hline
\end{tabular}}
\label{tab:rank}
\end{table*}

\section{Unlearning for Unintended Memorization}

The phenomenon where
LLMs tend to memorize certain training samples is called \textit{Memorization}. Carlini \textit{et al.} \cite{carlini2021extracting} extracted hundreds of GPT-2 training samples, including sensitive information such as address, email, phone, etc. The memorization can be troublesome if LLMs are trained on highly sensitive domain data, e.g. medical. Previous work studied the memorization of canaries and unlearned canaries from the training set of an LSTM model \cite{warnecke2021machine}.
However, the canaries can be viewed as outliers compared with the distribution of the original training set, which makes the removal easier.

To evaluate the effectiveness of the aforementioned unlearning methods against general memorization, we conduct two case studies on LLMs trained on a real-world medical dataset and an email dataset. 

\subsection{Datasets} 

\noindent \textbf{Medical Transcription(MT):}  MT dataset\footnote{https://www.kaggle.com/datasets/tboyle10/medicaltranscriptions} contains sample medical transcriptions for various medical specialties, including surgery, consultation, etc. There are 5k long medical transcriptions in total. We further split each transcription into shorter segments such that their lengths are below the input limit.\\

\noindent \textbf{Enron Email:}
This dataset contains approximately 500k emails generated by employees of the Enron Corporation\footnote{https://www.kaggle.com/datasets/wcukierski/enron-email-dataset}. It was obtained by the Federal Energy Regulatory Commission during its investigation of Enron's collapse.
There exist sensitive entities such as phone numbers and passwords in those emails. \\

\subsection{Training and Inference}

\noindent\textbf{Training.}
For each dataset mentioned above,
we first randomly select a subset of 100k samples and continue to train a pre-trained GPT-NEO-125M/OPT-125M on it for an extra 15 epochs.\\ 

\noindent \textbf{Inference.}
We randomly sample a batch of 128 texts containing more than 200 tokens from each dataset for the purpose of inference.
After applying an unlearning method to the trained LLM once,
we perform inference on the model using the prefixes of 10 tokens. 
Top-k \cite{li2020sampling} sampling is applied and each prefix is repeatedly inferred 10 times.

\subsection{Results}
Table \ref{table:unlearn-mem} displays how effective unlearning methods are at reducing memorization. Since the memorized samples are part of the real training set instead of canaries, the complete forgetting of them is more challenging. Although retraining can guarantee the complete removal, it takes 10X longer time than other methods. Among other unlearning baselines, Fisher Removal is the most effective, while Fisher Forgetting and gradient ascent are relatively weaker. 
Their performances are consistent with 
their efficacy scores in Table \ref{tab:rank}.

\subsection{Qualitative analyses} 
To understand the reason behind indelible memorizations, we manually inspect the training samples in the MT dataset. As shown in Table \ref{quali-mem}, these training samples have highly similar substrings. It has been proven that the rate at which LLMs regenerate training sequences is superlinearly related to a sequence's count in the training set \cite{kandpal2022deduplicating}. In order to handle these strong memorizations, data deduplication methods should be involved before the training stage. However, data deduplication falls in the category of pre-processing not unlearning.

\begin{table*}
\centering
\caption{Unlearning on MT and Enron Email Dataset}
\resizebox{1\textwidth}{!}{%
    \begin{tabular}{|c|cc|cc|cc|cc|}
    \hline
    Methods  & \multicolumn{4}{c|}{MT} & \multicolumn{4}{c|}{Enron} \\
    \cline{2-9}
    &\multicolumn{2}{c|}{NEO}   & \multicolumn{2}{c|}{OPT}
    &\multicolumn{2}{c|}{NEO}   & \multicolumn{2}{c|}{OPT} \\
    
   & \# of Mem  & Runtime (s) & \# of Mem & Runtime (s)
   & \# of Mem  & Runtime (s) & \# of Mem & Runtime (s)\\
      \hline
    No Unlearning &16 & - & 15 &-&20 & - & 22 &-\\
    \hline
   Retraining & 0&  $5.69 \times 10^4$ & 0& $5.81 \times 10^4$ 
   & 0 & $5.05 \times 10^4$ & 0 & $5.13 \times 10^4$\\
    \hline
   Finetuning & 16& $3.79 \times 10^3$ &15 & $3.87 \times 10^3$ & 20 & $3.42 \times 10^3$
    & 22 & $3.43 \times 10^3$  \\ \hline
  Gradient Ascent   &8 & $4.00 \times 10^0$ &7 & $7.00\times 10^0$  &9 & $2.30 \times 10^1$ &10 & $2.40 \times 10^1$\\
    \hline
  Fisher Removal   &4 & $1.57 \times 10^3$ &4 &$1.56 \times 10^3$ &5 & $1.10 \times 10^3$ &5 & $1.12 \times 10^3$  \\  \hline
 Fisher Forgetting  &8 & $1.57 \times 10^3$ & 7&  $1.56 \times 10^3$ & 8& $1.10 \times 10^3$  & 10 & $1.12 \times 10^3$ \\
    \hline
    \end{tabular}%
    }
    \label{table:unlearn-mem}
\end{table*}

\section{DP-SGD vs. Unlearning}
\begin{figure*}[!ht]
        \centering
        \begin{subfigure}[b]{0.45\textwidth}
            \centering
            \includegraphics[width=\textwidth]{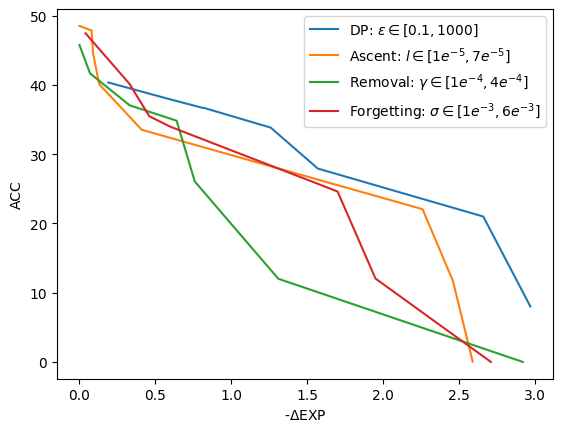}
            \caption[]%
             {{\small Lambada}}  
            \label{fig:dp-lambada}
        \end{subfigure}
        \hfill
        \begin{subfigure}[b]{0.46\textwidth}  
            \centering 
            \includegraphics[width=\textwidth]{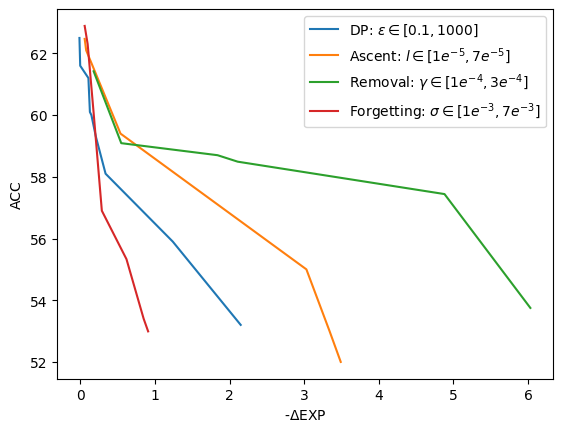}
            \caption[]%
            {{\small Piqa}}  
            \label{fig:dp-piqa}
        \end{subfigure}
    \caption{Privacy-utility trade-offs of DP-SGD and unlearning approaches on (a) Lambda and (b) Piqa Dataset. The model under evaluation is OPT-125M. }
    \label{fig:dp-trade-off}
    \end{figure*}

Differential privacy (DP) guarantees a bound of how much an individual sample can influence the output of a specific function. In the context of deep learning, DP-SGD \cite{abadi2016deep} is usually applied during the training phase. We aim to study to what extent a DP-SGD-trained LLM can  
protect the training set as well as compare DP-SGD against unlearning. 

\subsection{DP-SGD}
 We follow the standard definition in \cite{abadi2016deep}. A randomized mechanism $\mathcal{M}: \mathcal{D} \rightarrow \mathcal{R}$ with domain $\mathcal{D}$ and range $\mathcal{R}$ satisfies $(\epsilon, \delta)-$ differential privacy if any two adjacent inputs $d, d' \in \mathcal{D}$ for any subset of outputs $S \subseteq \mathcal{R}$ it holds that

$$\mathcal{P}[\mathcal{M}(d) \in S] \leq e^{\epsilon} \mathcal{P}[\mathcal{M}(d') \in S] +\delta$$

To investigate whether DP-SGD offers guaranteed privacy in practice,  
we audit DP-SGD under varying privacy budget $\epsilon$ values. Specifically, we observe the exposure scores of the $128$-sample subset by querying the DP-SGD trained models. Finally, the privacy-utility trade-off is obtained.

\subsection{Privacy-Utility Trade-off}
Likewise, we apply the previous unlearning methods to remove the same $128$ samples from the model trained without DP-SGD and report the final exposure scores. 
By varying the coefficients (in Table \ref{tab:para_def}), we can generate a privacy-utility trade-off for each method.
To this end, the trade-offs of DP-SGD as well as the unlearning approaches are summarized in Figure \ref{fig:dp-trade-off}. 

First, DP-SGD does not guarantee an equally optimal/suboptimal trade-off on different datasets.
Although DP almost dominates other methods on the Lambada dataset, its behavior is not robust on the Piqa dataset. Second, DP-SGD introduces additional computational costs into the trivial training process, making the development of LLMs more expensive. Therefore, DP-SGD may serve as a general solution to protect training samples, but it
cannot play the same role of unlearning. 

\section{Related Work}
\subsection{Cost of LLMs}
The capacities of LLMs are usually accompanied by tremendous GPU hours. Recently, Touvron \textit{et al.} \cite{touvron2023llama} revealed that training a LLaMA-7B model requires 82432 hours on Nvidia A100 80GB GPU, not to mention larger models such as OPT-175B \cite{zhang2022opt}. Given the enormous amount of resources required for a single training session, erasing target data by training from scratch is impractical and economically prohibitive for LLMs. For example, retraining a model like OPT-175B would entail not only substantial computational power but also significant time and energy costs, translating into substantial financial burdens.

For service providers, efficient unlearning methods
maybe a practical solution 
to significantly reduce computational overhead while still observing the data regulations.

\subsection{Machine Unlearning}

Cao \textit{et. al} \cite{Cao2015TowardsMS} firstly introduced machine unlearning for statistical query learning. 
They proposed a layer of summations between the learning algorithm and the training data. The learning algorithm relied only on these summations, each of which was the sum of some transformations of the training data samples. The summations were computed during the training phase and used to update the model during the unlearning phase. To forget a specific data sample, the approach simply updated a small number of summations that depended on that sample and then updated the model by resuming the learning algorithm for a few more iterations.

Certified data removal \cite{guo2019certified} was later proposed, which provided a theoretical guarantee for removal from all convex models without the need to retrain them from scratch. The key method involved using Newton update to modify model parameters and reduce the influence of deleted data points. This process was complemented by masking any residual information with random perturbations, ensuring that removed data cannot be inferred. The method was shown to be practical in various settings, providing a balance between data removal efficacy and model performance.

As for more general removal that is compatible with deep neural networks,
Bourtoule \textit{et. al} \cite{bourtoule2021machine}  proposed a sharding approach by diving the training data into multiple disjoint shards and training a model on each shard.
When the request for unlearning arrived, only the constituent model associated with the specific data shard needed to be retrained. While effectively reducing the computational overhead of unlearning in machine learning models without significantly impacting the accuracy for simpler learning tasks, it introduces several challenges while working with LLMs. Assuming that each shard adequately represents the overall data distribution is not realistic. Moreover, training multiple models on large shards while tuning the hyperparameter for each constituent model also demands significant computational resources. This involves not just the processing power but also memory requirements, as each constituent model needs to be stored, trained, and optimized separately before integration.

More recently, the unlearning of LLMs has attracted increasing attention.
Yao \textit{et al.} \cite{yao2023llmunlearn} applied gradient ascent to unlearn undesirable behaviors in LLMs by aligning the model using negative examples. 
Wang et al. \cite{wang-etal-2023-kga} presented a novel approach that focused on aligning the knowledge gap between models trained on different datasets,  suggesting that unlearned models should treat
the deleted data as unseen data. 
However, acquiring high-quality data from the same distribution can be particularly challenging, especially when the availability of such data is limited. 
In addition, we aim to generally remove the effects of unlearning subsets, rather than focusing on aligning LLMs with human preferences or knowledge gaps.

 Chen and Yang \cite{chen-yang-2023-unlearn} introduced an efficient unlearning framework designed to update LLMs effectively without the need for full retraining. This was achieved through the integration of lightweight unlearning layers into transformers, which were specifically trained on a set of data designated for forgetting, enabling it to effectively discard specific knowledge.  
 Elden \textit{et al.} \cite{eldan2023s} proposed to replace sensitive tokens of Harry Potter-related text
with generic counterparts and finetune the LLMs on the modified samples. Nevertheless, the replacement process is non-trivial and does not easily generalize to other domains.
Jang \textit{et al.} \cite{jang2022knowledge} showed that applying gradient descent directly can make LLMs forget about target samples, while no guarantee is provided for model utility.

 The aforementioned work has only explored the unlearning task with first-order information, which is reasonable due to its extreme efficiency. However, most of them either require data engineering \cite{yao2023llmunlearn, wang-etal-2023-kga, eldan2023s} or specific architectural design \cite{chen-yang-2023-unlearn}, which limits the generalizability. We explore novel unlearning approaches for LLMs relying on second-order information. We show that Hessian can be estimated with affordable computational cost and our methods demonstrate superior robustness with respect to gradient ascent.

\begin{figure*}[!htb]

    \begin{subfigure}{0.3\textwidth}
        \includegraphics[width=\textwidth]{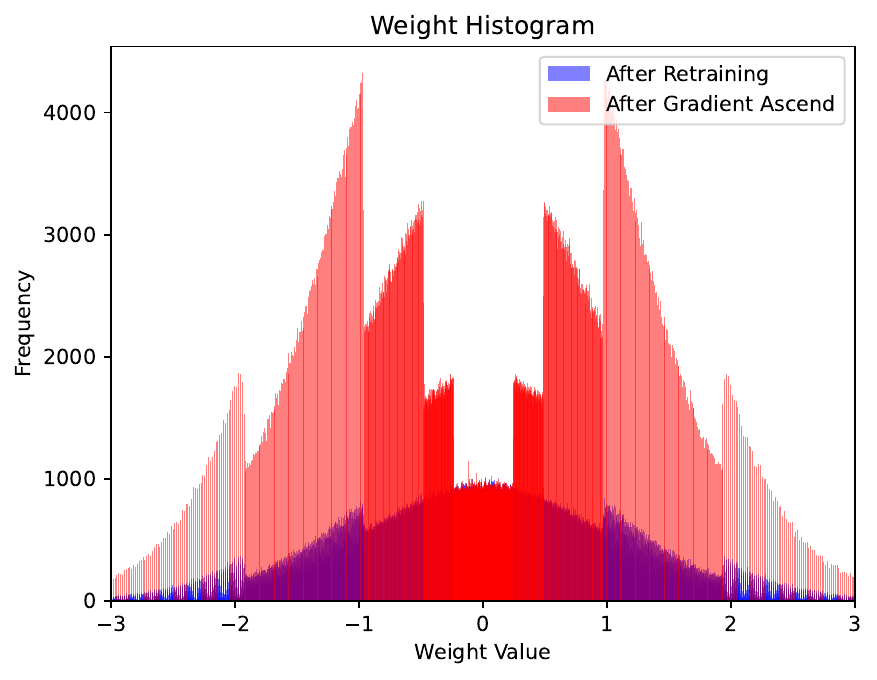}
        \caption[]%
        {{\small KL = 1176792}}
        \label{fig:arc_gpt-neo_2_256_ascend}
    \end{subfigure}
    \begin{subfigure}{0.3\textwidth}
        \includegraphics[width=\textwidth]{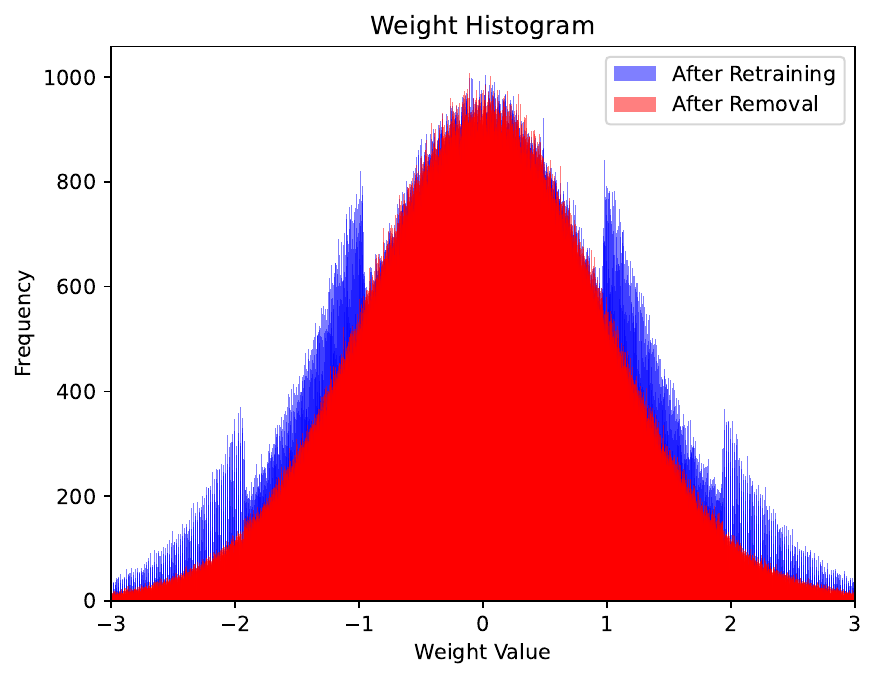}
        \caption[]%
        {{\small KL = 425}}
        \label{fig:arc_gpt-neo_2_256_removal}
    \end{subfigure}
    \begin{subfigure}{0.3\textwidth}
        \includegraphics[width=\textwidth]{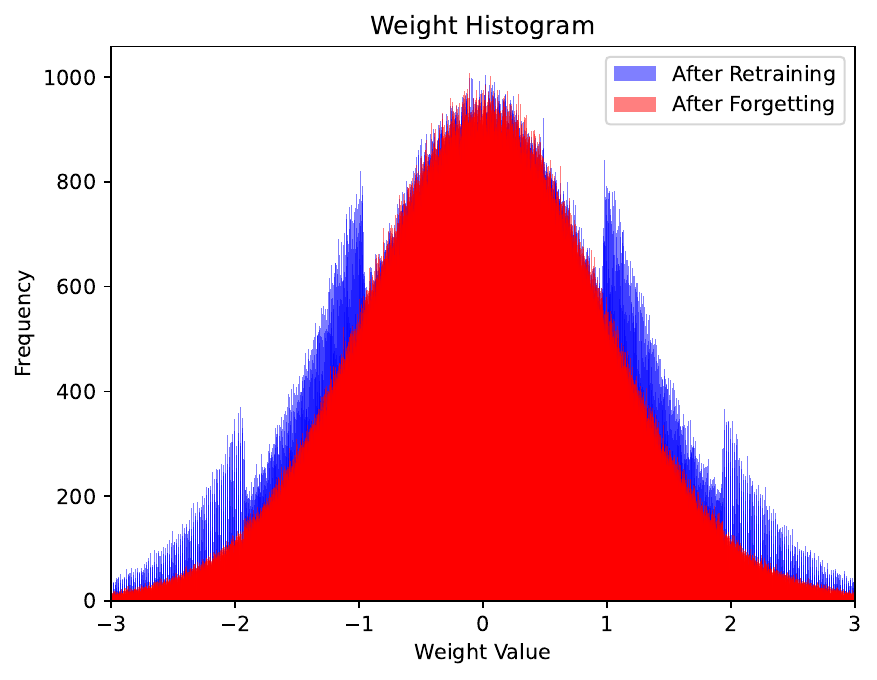}
        \caption[]%
        {{\small KL = 4174}}
        \label{fig:arc_gpt-neo_2_256_forget}
    \end{subfigure}
    \caption{Weights distribution of the last layer of GPT-NEO (125M) fine-tuned on ARC dataset. \# of bins for plotting the histograms is set to 40k. (a), (b) and (c) display the distribution after applying Gradient Ascent, Fisher Removal, and Fisher Forgetting respectively.
    }
    \label{weight_dist}
\end{figure*}%

\section{Discussions}

\subsection{Interpreting Unlearning via Weight Distributions}

Recall that the distribution of the model parameters after an optimal unlearning algorithm should be indistinguishable from a model that has never seen the unlearning subset (eq. \ref{KL-dif}). Specifically, the KL quantity $KL(\mathcal{P}(S_1(\theta,D ))||\mathcal{P}(S_2(\theta, D^+))) $ should be
small (or close to zero in ideal setting). To better illustrate the effects of unlearning, we visualize the weight of the last layer in GPT-NEO 125M after performing different unlearning algorithms in Figure \ref{weight_dist}.

First, it is noticeable that the weight distribution after applying the gradient ascent deviates far from that of retraining, leading to a large KL divergence of 1176792.
However, our second-order-based methods achieve unlearning without pushing the weight distribution too far from retraining. The quantitative KL divergence values by Fisher Removal and Fisher Forgetting are merely 425 and 4174 respectively,  which are consistent with the histograms.

\subsection{Extended Unlearning Cycles}
\label{extended-unlearning}
We have discussed the property of Fisher Forgetting in Section \ref{sec:fisher-forgetting}, here we compare the performance of each baseline when unlearning cycles are further extended.

\begin{figure}[!htb]
        \centering
        \begin{subfigure}[b]{0.24\textwidth}
            \centering
            \includegraphics[width=\textwidth]{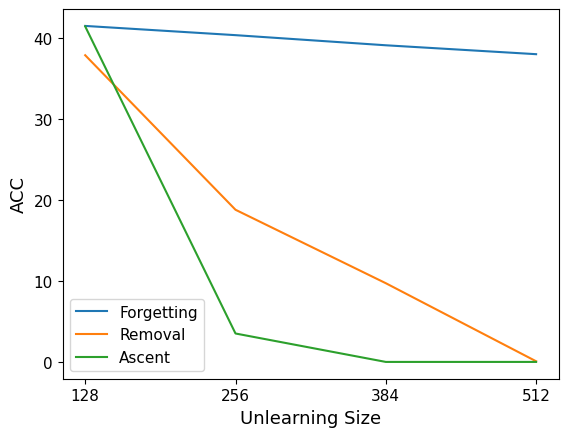}
            \caption[]%
            {{\small Lambada  }}    
            \label{fig:extend-lambada}
        \end{subfigure}
        ~
        \begin{subfigure}[b]{0.24\textwidth}  
            \centering 
            \includegraphics[width=\textwidth]{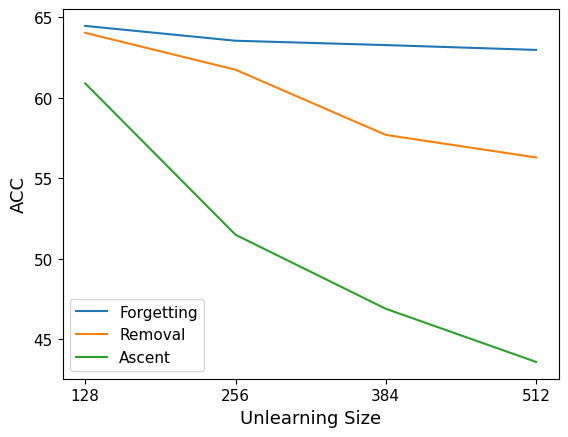}
            \caption[]%
            {{\small Piqa }}    
            \label{fig:extend-piqa}
        \end{subfigure}
    \caption{Model utility curves after extended unlearning cycles. Both (a) and (b) are generated by GPT-NEO-125M model. }
    \label{fig:extending_unlearing_cycle}
    \end{figure}

The GPT-NEO-125M performance curves after consecutive unlearning cycles are displayed in Figure \ref{fig:extending_unlearing_cycle}. As indicated by the theoretical analysis, Fisher Forgetting maintains the model utility robustly on both the Lambada and Piqa datasets. However, it is not guaranteed that other methods will retain a similar level of utility.
 Specifically, we observe that gradient ascent causes model performance to drop rapidly as the unlearning subset grows in both scenarios. 

\subsection{Onion Effect}

Carlini \textit{et al.} \cite{carlini2022privacy} revealed and analyzed an \textit{Onion Effect} of memorization: removing the ``layer'' of outlier points that are most vulnerable to a privacy attack exposes a new layer of previously safe points to the same attack. In the context of machine unlearning, we explore how the removal of certain data samples impacts the rest of the training data.

\begin{figure}[!htb]
\includegraphics[width=8cm]{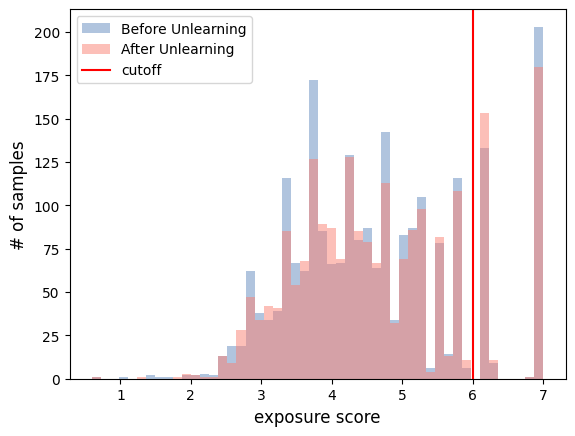}
\caption{Onion effect on GPT-NEO-125M. In the ARC dataset, we set the cutoff threshold as $6$ and remove all the samples with higher exposure scores. After retraining the model, some left samples have their exposure scores increase past the threshold.}
\label{fig:onion effect}
\end{figure}

We conduct a control experiment on the ARC dataset using GPT-NEO-125M. Specifically, we first obtain the exposure score of each training sample by querying the model before unlearning. Second, samples with exposure scores higher than $6$ are removed. Finally, we retrain the model using the modified training dataset and recalculate the exposure scores. As shown in Figure \ref{fig:onion effect}, some initially ``safe'' samples become more exposed after the most exposed samples are removed, which validates the existence of the onion effect.

The results imply that machine unlearning may contradict the goal in the case of eliminating the most vulnerable samples. However, in general, in cases where we assume the unlearning subset is randomly sampled, no significant shift of exposure distributions is observed after unlearning. Figure \ref{fig:random-onion} demonstrates the stability of exposure distributions after unlearning.

\section{Limitations \& Future Work}

\subsection{Limitations}

\subsubsection{Calculating Hessian is Relatively Costly\\}

 We have shown that the time complexity of estimating Hessian can be reduced to the product between the size of the unlearning subset and the number of recursions. However, it is still relatively costly compared to the calculation of first-order information. For LLMs, extra GPU space may be required to store the Hessian matrices. Both the time complexity and space complexity could be improved by optimizing the current estimation formula.

\subsubsection{Erasing Unintended Memorization is Challenging\\}
Although previous work showed that canaries inserted into the training set can be easily erased from an LSTM model 
\cite{warnecke2021machine}, it should be considered a special case of memorization since canaries are outliers in the regular training set. Our evaluation of general memorization reveals that even Fisher Removal cannot guarantee complete forgetting in one shot. However, repeatedly applying the unlearning algorithm may compromise the model's utility. 
Therefore, it requires further investigation such as data duplication \cite{kandpal2022deduplicating} to address indelible memorization.

\subsection{Future Work}

\subsubsection{Towards Better Privacy-Utility Trade-off\\}
If we consider that retraining represents the optimal trade-off
between privacy and utility, Fisher Removal is biased toward the side of privacy, while Fisher Forgetting favors the side of utility. It still remains challenging to design an unlearning algorithm that strongly guarantees both properties. Future work can focus on improving the privacy-utility trade-off of unlearning. One possible direction is combining different unlearning strategies such as our Fisher Removal and Fisher Forgetting.

\subsubsection{Robustness against Extended Unlearning Cycles\\}
In practice, an unlearning algorithm should preserve the utility of the LLM even after multiple unlearning procedures. Otherwise, the service provider may still resort to retraining frequently due to the loss of utility, which contradicts the purpose of unlearning. In our evaluation, only Fisher Forgetting possesses this desired robustness. However, this property is imperative for any unlearning algorithm to be useful in real-world applications. Future work should focus on developing and refining unlearning methods that can reliably preserve the utility of LLMs across multiple unlearning cycles. 

\subsubsection{Unlearning Larger LLMs \& Better Evaluation Metrics\\}
Due to the hardware constraints, we have mainly performed experiments on GPT-NEO/OPT models up to 2.7B. Since our unlearning methods are model-agnostic, we encourage future work to explore the unlearning process of larger LLMs such as LLama2 7B and Falcon 7B. 
In addition, we largely follow the previous work to measure how much an LLM still remembers the unlearning subset by exposure score. However, the setup of subspace and the process of iterating through all the samples can be time-consuming. 
Future work might investigate more efficient/elegant metrics to audit the outcome of unlearning.

\section{Conclusion}
In this paper, we propose two novel unlearning algorithms for large language models, namely Fisher Removal and Fisher Forgetting. Although they are both derived from Newton update, Fisher Removal provides a stronger guarantee for the erasure of the unlearning subset compared with gradient ascent. Fisher Forgetting is designed to preserve the utility of LLMs against multiple unlearning processes.
Through a comprehensive evaluation of four NLP datasets, we demonstrate that second-order information can make the unlearning outcomes more robust compared to using only first-order information.

We further reveal that unlearning can mitigate unintended memorizations, but there still exist indelible memorizations that require additional prevention approaches like data deduplication. Finally, we uncover the relationship between unlearning and DP-SGD through the privacy-utility trade-off. The results suggest that DP-SGD does not always guarantee an equally optimal/sub-optimal trade-off across different datasets, which implies that it cannot play the same role of unlearning. Instead, DP-SGD may serve as a general solution to protect training samples.

For future work, we suggest improving existing unlearning methods towards better privacy-utility trade-offs as well as robustness against multiple unlearning cycles. Our unlearning algorithms can also benefit from adopting more efficient Hessian approximation formulas.

\bibliographystyle{ACM-Reference-Format}
\bibliography{references}

\appendix

\section{Appendix}
\subsection{Proof of Newton Update}
\label{proof_1}
Let the loss be $L_D(\theta)  = L_{D^+}(\theta) + L_{D^-}(\theta)$ and assume that the loss function $L(\cdot)$ is quadratic. The optimization algorithm $\mathcal{A}_t(D, \epsilon )$ is given by the gradient information at time step $t$ with random initialization. Consider the following unlearning process:

\begin{equation}
    h(\theta) = \theta + e^{- H_1t}e^{H_2 t} V_1 +  e^{-H_1 t}(V_1 - V_2) - V_2
\end{equation}

Where $H_1 = \nabla^2 L_{D}(\theta), H_2 = \nabla^2 L_{D^+}(\theta),V_1 = H^{-1}_1 \nabla L_D(\theta)$, and $V_2 = H^{-1}_{2} \nabla L_{D^+}(\theta)$.
Then $h(\theta)$ satisfies $h (\mathcal{A}_t(D, \epsilon )) = \mathcal{A}_t(D, \epsilon )$ for 
all random initializations and all times $t$. In particular, let $S_1(\theta, D) = h(\theta)$, which removes all the information of $D^-$:

$$KL(\mathcal{P}(S_1(\theta, D))||\mathcal{P}(\mathcal{A}(D^+, \epsilon)) = 0$$

Note that when $t \rightarrow \infty$, that is, when the optimization algorithm has converged,
the unlearning process reduces to the simple Newton update.

\begin{table*}[t]
    \centering
    \caption{Unlearning Results on Lambada Dataset \\  }
\resizebox{\textwidth}{!}{ 
 \begin{tabular}{cc|cccc|cccc}
    \hline
    \textbf{Parameter}  & \textbf{Unlearning} &\multicolumn{4}{c|}{\textbf{128 samples}} & \multicolumn{4}{c}{\textbf{256 samples}} 
      \\ \cline{3-10}
      & & \multicolumn{2}{c}{NEO} & \multicolumn{2}{c|}{OPT}&
      \multicolumn{2}{c}{NEO} & \multicolumn{2}{c}{OPT}\\
      \hline
        & &  $\Delta_{ACC}$   & $\Delta_{EXP}$  & $\Delta_{ACC}$  & $\Delta_{EXP}$ &  $\Delta_{ACC}$   & $\Delta_{EXP}$  & $\Delta_{ACC}$  & $\Delta_{EXP}$  \\ \hline

       \multirow{6}{*}{125M} 

       &  Retraining & -1.61 & -0.30  & -4.85 &-0.43  & +0.85 &-0.24& -1.34 & -0.48\\
       
       & Finetuning & -1.98 & -0.27 &\cellcolor{blue!10}-2.48  &\cellcolor{gray!25} +0.33 & +0.58   &  \cellcolor{gray!25}+0.38 &
       -1.84&\cellcolor{gray!25} +0.22\\
       
      & Gradient Ascent &\cellcolor{blue!10} -0.74 &  -0.12 &  -8.79& -0.13 & \cellcolor{gray!25}-38.66 & \cellcolor{blue!10}-1.99 
      & \cellcolor{gray!25}-48.90
      & \cellcolor{gray!25}+0.35\\
      & Fisher Removal  & -4.30  &\cellcolor{blue!10} -0.31 & \cellcolor{gray!25}-14.05 & \cellcolor{blue!10}-0.64  &\cellcolor{gray!25}-23.23 &\cellcolor{blue!10}-0.99
      &\cellcolor{gray!25}-48.90 & \cellcolor{blue!10}-1.68\\
      & Fisher Forgetting & \cellcolor{blue!10}-0.68  & -0.14 &\cellcolor{blue!10} -1.43 &-0.14  &-1.82 & -0.04 & -4.04 & -0.09\\
        \hline
      \multirow{6}{*}{1.3B} 
      &  Retraining &-1.20 &-0.49 & +1.89&-0.76 &-0.72 &-0.81&+1.96 &-0.46\\
      & Finetuning &\cellcolor{blue!10}-0.89 & -0.25  & +1.26& \cellcolor{gray!25}+0.52 &\cellcolor{blue!10} -0.68 &\cellcolor{gray!25} +0.27
      & + 1.13 & \cellcolor{gray!25}+0.38\\

      & Gradient Ascent  & \cellcolor{blue!10}-0.31& -0.28&\cellcolor{gray!25} -28.74&-0.23 &
\cellcolor{gray!25}-58.43&\cellcolor{gray!25}+0.55  &\cellcolor{gray!25} -58.72 & \cellcolor{blue!10}-4.16\\
      & Fisher Removal &-2.48&\cellcolor{blue!10}-0.50 &-2.72 &-0.48 & \cellcolor{blue!10}+0.47& \cellcolor{blue!10}-1.81 
      & \cellcolor{gray!25}-24.78 &\cellcolor{blue!10}-1.51\\
      & Fisher Forgetting &  \cellcolor{blue!10}-0.60 &-0.05 &-1.30 &-0.13 & \cellcolor{blue!10}-0.52&\cellcolor{gray!25}+0.05&
      -4.28 & -0.18\\
      \hline
      
      \multirow{6}{*}{2.7B} 
      &  Retraining &-0.86 &-1.10 &-1.12 &-0.29 &+1.32 & -1.24 & -0.27 &-1.25\\
      & Finetuning &-0.92& \cellcolor{gray!25} +0.16& \cellcolor{blue!10} -0.33&\cellcolor{gray!25}+0.10 &-2.60 &\cellcolor{gray!25} +0.17
      &-2.71 &\cellcolor{gray!25}+0.10 \\
      
      & Gradient Ascent &\cellcolor{blue!10}+1.98 & -0.41& \cellcolor{gray!25}-61.63& \cellcolor{blue!10}-3.76& \cellcolor{gray!25}-59.60&\cellcolor{blue!10}-1.76
      &\cellcolor{gray!25}-61.63& \cellcolor{blue!10}-4.44\\
      & Fisher Removal  &-2.97 & \cellcolor{blue!10}-1.10&-2.61& -0.25 &\cellcolor{gray!25}-56.79 &\cellcolor{blue!10}-3.01 &\cellcolor{gray!25}-60.62 & -1.18\\
      & Fisher Forgetting & -1.07& -0.05
      &\cellcolor{blue!10} -0.60&-0.05 &-2.37 & -0.06 & -3.22 & -0.11\\

      \hline  
    \end{tabular} 
   }
    
    \label{tab:Unlearning_Lambada}
\end{table*}

\begin{table*}
    \centering
    \caption{Indelible Memorizations in MT dataset}
    \begin{tabular}{|p{16cm}|}
    
       \hline 
           The incision site was closed with 7-0 nylon.,
          \textcolor{blue}{The patient tolerated the procedure well and 
          was transferred to the recovery room in stable condition} with Foley catheter in position. \\
          
      \hline
       She was subsequently consented for a laparoscopic appendectomy.,DESCRIPTION OF PROCEDURE: , After informed consent was obtained, \textcolor{blue}{the patient was brought to the operating room, placed supine on the table.}\\
      \hline 
      PROCEDURE PERFORMED: , Modified radical mastectomy.,ANESTHESIA: , General endotracheal tube.,PROCEDURE:  ,After informed consent was obtained, \textcolor{blue}{
      the patient was brought to the operative suite and placed supine on the operating room table.}\\
      \hline
    \end{tabular}
    \label{quali-mem}
\end{table*}

\begin{figure*}
    \includegraphics[width=\textwidth]{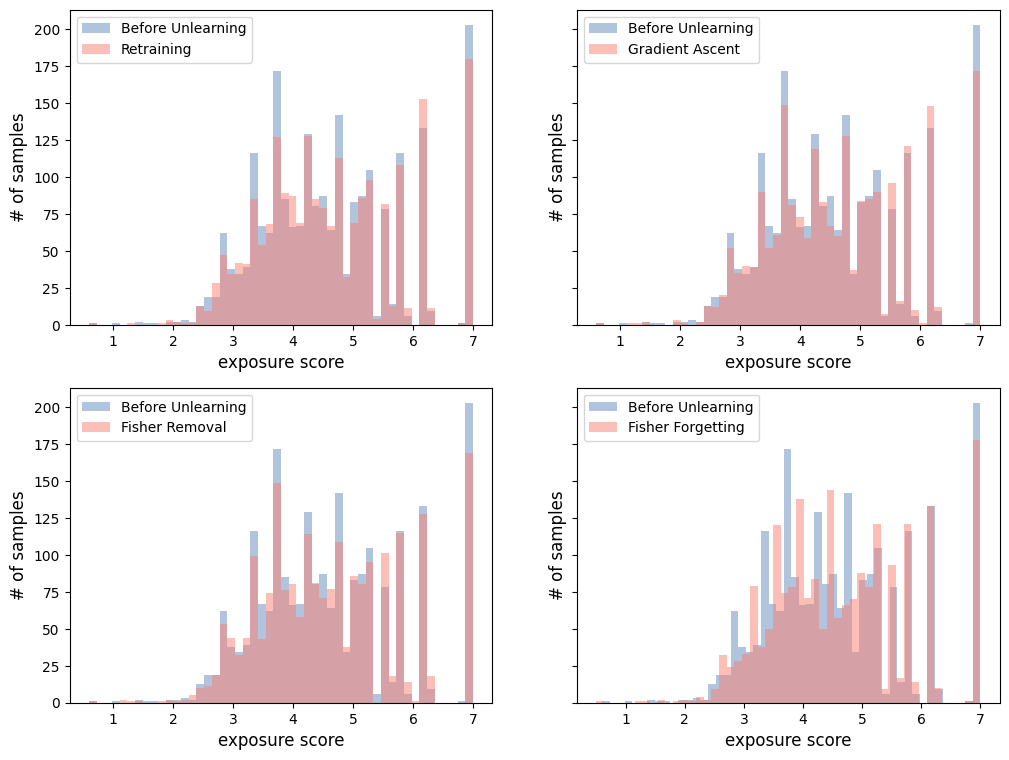}
    \caption{The distributions of exposure scores on ARC before/after unlearning.}
    \label{fig:random-onion}
\end{figure*}

\end{document}